\documentclass[conference]{IEEEtran}

\usepackage{amsmath,amssymb,amsfonts}
\usepackage[plainruled,vlined]{algorithm2e}
\usepackage{graphicx}
\usepackage{xcolor}

\usepackage{newtxtext}
\usepackage{newtxmath}
\usepackage[varqu,varl]{zi4}
\usepackage{gensymb}
\usepackage{textcomp}
\usepackage[mathb]{mathabx}
\usepackage{bbm}
\usepackage{xfrac}
\usepackage{soul}
\usepackage{enumitem}
\setlist{nolistsep}
\usepackage{tikz}

\usepackage{fancyhdr}
\usepackage[normalem]{ulem}
\usepackage[hyphens]{url}
\usepackage[sort,nocompress]{cite}
\usepackage[final]{microtype}
\usepackage[keeplastbox]{flushend}
\usepackage{hyperref}
\hypersetup{bookmarks=true,letterpaper=true,colorlinks,linkcolor={blue!50!black},citecolor={blue!50!black},urlcolor={blue!50!black}}

\usepackage{float}
\usepackage{booktabs}
\usepackage{subcaption}

\newcommand{\ours}{Procrustes}
\newcommand{\conv}{\textsc{conv}}
\newcommand{\fc}{\textsc{fc}}

\title{\ours{}: a Dataflow and Accelerator\\ for Sparse Deep Neural Network Training}
\author{
\IEEEauthorblockN{Dingqing Yang\IEEEauthorrefmark{2}, Amin Ghasemazar\IEEEauthorrefmark{2}\IEEEauthorrefmark{1}, Xiaowei Ren\IEEEauthorrefmark{2}\IEEEauthorrefmark{1}, Maximilian Golub\IEEEauthorrefmark{3}, Guy Lemieux\IEEEauthorrefmark{2}, and Mieszko Lis\IEEEauthorrefmark{2}}
\IEEEauthorblockA{
{
\IEEEauthorrefmark{2}The University of British Columbia  \qquad\qquad \IEEEauthorrefmark{3}Microsoft Corporation}\\
\texttt{\{dingqingy,aming,xiaowei\}@ece.ubc.ca, magolub@microsoft.com, \{lemieux,mieszko\}@ece.ubc.ca}}
}

\newcommand{\pubtag}{\small\emph{Appears in the Proceedings of the 53$^\mathit{rd}$ IEEE/ACM International Symposium on Microarchitecture (MICRO~2020)}}

\begin{document}
\maketitle
\thispagestyle{fancy}
\pagestyle{fancy}
\renewcommand{\headrulewidth}{0pt}
\chead{\pubtag}

\begin{abstract}

The success of DNN pruning has led to the development of energy-efficient inference accelerators that support pruned models with sparse weight and activation tensors. Because the memory layouts and dataflows in these architectures are optimized for the access patterns during \emph{inference,} however, they do not efficiently support the emerging sparse \emph{training} techniques.

In this paper, we demonstrate (a)~that accelerating sparse training requires a co-design approach where algorithms are adapted to suit the constraints of hardware, and (b)~that hardware for sparse DNN training must tackle constraints that do not arise in inference accelerators. As proof of concept, we adapt a sparse training algorithm to be amenable to hardware acceleration; we then develop dataflow, data layout, and load-balancing techniques to accelerate it.

The resulting system is a sparse DNN training accelerator that produces pruned models with the same accuracy as dense models without first training, then pruning, and finally retraining, a dense model. Compared to training the equivalent unpruned models using a state-of-the-art DNN accelerator without sparse training support, \ours{} consumes up to 3.26$\times$ less energy and offers up to 4$\times$ speedup across a range of models, while pruning weights by an order of magnitude and maintaining unpruned accuracy.

\end{abstract}

\section{Introduction}

\renewcommand{\thefootnote}{\fnsymbol{footnote}}
\footnotetext{\IEEEauthorrefmark{1}These authors contributed equally.}
\footnotetext{\IEEEauthorrefmark{3}Work done while the author was with the University of British Columbia.}
\addtocounter{footnote}{1}

\noindent Deep neural networks are known to be vastly overparameterized: pruning techniques can typically reduce the weight count by an order of magnitude~\cite[etc.]{lecun_optimal_1990,hassibi1993second,han2015learning,li2016pruning,handeep2016,yang2017designing,luo2017thinet}. This sparsity comes at the cost of irregular memory accesses and computation patterns, and several accelerators have been proposed to enable efficient inference on sparse models~\cite[etc.]{han2016eie,zhang2016cambriconx,parashar2017scnn,chen2019eyerissv2,gondimalla2019sparten}.

None of these approaches were, however, designed for energy-efficient \emph{training}. This is because they target a context where pruning occurs \emph{after} training: a model is first trained with the full parameter set, then pruned, and finally re-trained to recover accuracy~\cite{han2015learning}. While this saves energy at inference time, training the pruned network takes \emph{more} time and energy than training an equivalent dense network to the same accuracy. Skipping the pre-training step is not an option: even if oracular knowledge of the pruned model connectivity is assumed, training the pruned model from scratch sacrifices accuracy compared to the original network~\cite{han2015learning,li2016pruning}.

Still, the very existence of pruned networks suggests that it must be possible to somehow train them. Recent work has demonstrated that a model pruned by an order of magnitude can be trained \emph{provided} that the initialization for the unpruned subset of weights is preserved~\cite{frankle2019lottery}; this can be achieved either by dynamically selecting the most productive gradient subspace~\cite{golub2019dropback} or by iteratively increasing sparsity~\cite{mostafa2019parameter,zhang2019eager}.

\begin{figure}[t!]
\centering
\includegraphics{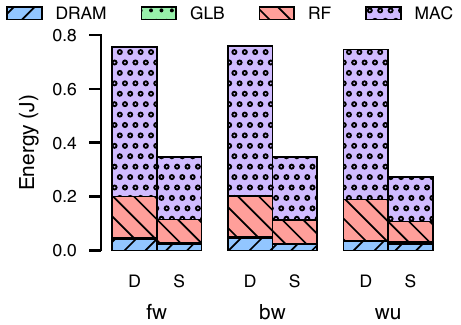}~\includegraphics{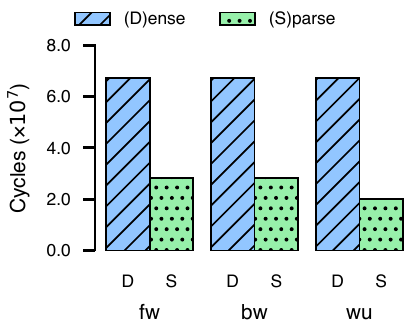}
\vspace{-3ex}
\caption{Potential training energy savings and speedup from ideally leveraging all weight sparsity (here, 5$\times$) while training VGG-S (15M weights) to convergence with Dropback~\cite{golub2019dropback}. fw/bw/wu = forward/backward/weight-update phases.}
\vspace{-2ex}
\label{fig:opportunity}
\end{figure}

Ideally, such sparse-from-scratch training can offer significant savings. \autoref{fig:opportunity} shows this for VGG-S~\cite{zagoruyko_torch_2015} pruned $5\times$ (15M$\rightarrow$3M weights) using the Dropback algorithm~\cite{golub2019dropback}, in an idealized 16$\times$16 PEs training system where (i)~sparsity is evenly distributed within each layer so all PEs receive the same workload (i.e., perfect load balancing), and (ii)~sparse weights are stored in an idealized compressed format with no overhead, and (iii)~retained weights selection is instant and cost-free (see \autoref{sec:methods} for setup details). While the exact improvement varies with the geometry and sparsity of each layer, leveraging $5\times$ sparsity can yield up to $2.6\times$ speedup with $2.3\times$ less energy consumption over the entire network.

In practice, however, none of the existing sparse training methods can reach this potential. Most~\cite{frankle2019lottery,zhang2019eager,golub2019dropback} require sorting all weights to determine the parameters to retain; with weight counts in the tens of millions, sorting is an expensive proposition. Several~\cite{mostafa2019parameter,zhang2019eager} achieve only small pruning factors and suffer accuracy loss. Some~\cite{frankle2019lottery,zhang2019eager} prune the model very gradually; this implies (i)~no peak memory footprint reduction, (ii)~mediocre energy savings because the average sparsity is low during most of the training process, and (iii)~the need to support two weight storage formats (dense and sparse) and switch formats mid-way during training. The remaining technique~\cite{golub2019dropback} maintains the target weight sparsity throughout training, but gives up computation sparsity --- a significant drawback for training, where weights are usually 32-bit floating-point numbers that are energetically expensive to multiply.

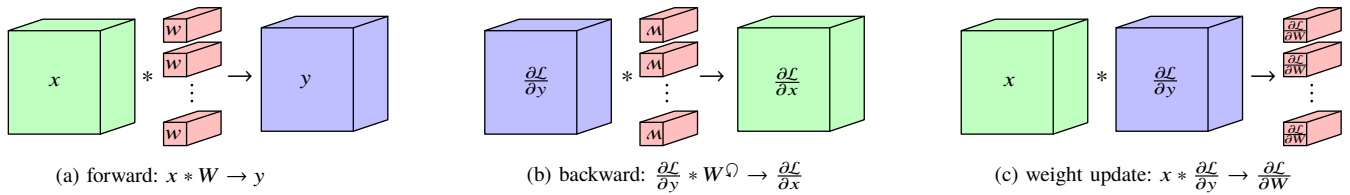
\begin{figure*}[t]
\center
\begin{tikzpicture}
\footnotesize
\node (img) {\includegraphics{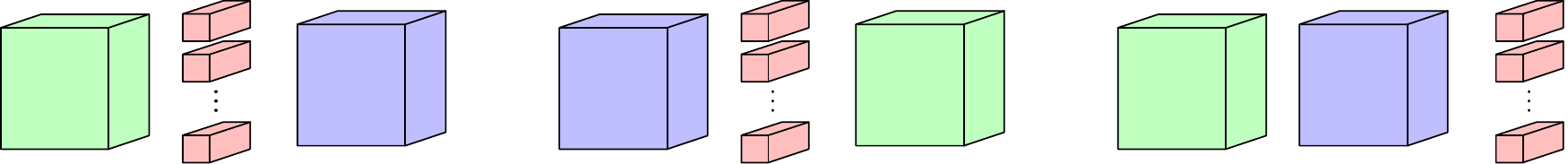}};
\node at (-235pt,-2pt){$x$};
\node at (-140pt,-2pt){$y$};
\node at (-190pt,17.5pt){$w$};
\node at (-190pt,4.5pt){$w$};
\node at (-190pt,-21.75pt){$w$};
\node at (-200pt,0pt){\normalsize$\ast$};
\node at (-165pt,0pt){\normalsize$\rightarrow$};
\node at (-195pt,-38pt){(a)~forward: $x \ast W \rightarrow y$};
\node at (-53pt,-2pt){$\frac{\partial\!\mathcal{L}}{\partial y}$};
\node at (42pt,-2pt){$\frac{\partial\!\mathcal{L}}{\partial x}$};
\node[rotate=180] at (-8.5pt,17.5pt){$w$};
\node[rotate=180] at (-8.5pt,4.5pt){$w$};
\node[rotate=180] at (-8.5pt,-21.75pt){$w$};
\node at (-18.5pt,0pt){\normalsize$\ast$};
\node at (13.5pt,0pt){\normalsize$\rightarrow$};
\node at (-4pt,-38pt){(b)~backward: $\frac{\partial\!\mathcal{L}}{\partial y} \ast W^{\uptodownarrow} \rightarrow \frac{\partial\!\mathcal{L}}{\partial x}$};

\node at (126pt,-2pt){$x$};
\node at (185pt,-2pt){$\frac{\partial\!\mathcal{L}}{\partial y}$};
\node at (234.25pt,17.5pt){\tiny \scalebox{0.8}{$\frac{\partial\!\mathcal{L}}{\partial W}$}};
\node at (234.25pt,4.5pt){\tiny \scalebox{0.8}{$\frac{\partial\!\mathcal{L}}{\partial W}$}};
\node at (234.25pt,-21.75pt){\tiny \scalebox{0.8}{$\frac{\partial\!\mathcal{L}}{\partial W}$}};
\node at (161pt,0pt){\normalsize$\ast$};
\node at (222pt,0pt){\normalsize$\rightarrow$};
\node at (177pt,-38pt){(c)~weight update: $x \ast \frac{\partial\!\mathcal{L}}{\partial y} \rightarrow \frac{\partial\!\mathcal{L}}{\partial W}$};
\end{tikzpicture}
\vspace{-1ex}
\caption{CNN training consists of (a)~the forward pass, (b)~the backward pass, and (c)~the weight update pass; minibatch size adds a fourth dimension to the activations. Weights are accessed in different order during the forward and backward passes. Training \fc{} layers is similar but uses multiplication instead of convolution and $W^\top$ instead of  $W^{\uptodownarrow}$ in the backward pass. $\mathcal{L}$ = loss; $x$ = iacts; $y$ = oacts; $W$ = weights; $\uptodownarrow$~=~$180\degree$ filter-wise rotation.}
\label{fig:training-stages}
\end{figure*}

In addition, existing accelerators that support sparse inference are inadequate for sparse training. Weights are represented in formats that directly correspond to the dataflow being used~\cite[etc.]{han2016eie,zhang2016cambriconx,parashar2017scnn,chen2019eyerissv2,gondimalla2019sparten}; this works well when weights are always accessed in the same order during inference, but does not support the different weight access patterns that arise in different phases of training (see \autoref{sec:background:training}). Accelerators that perform load balancing (e.g., Sparten~\cite{gondimalla2019sparten}) do this in software as a preprocessing step; this works for inference where weight sparsity is static, but not for training where weight sparsity changes dynamically. Finally, recent proposals like SCNN~\cite{parashar2017scnn} and Sparten~\cite{gondimalla2019sparten} use complex hardware to exploit two-sided sparsity (i.e., both weight and activation sparsity); this can be leveraged during the forward-pass phase of training, but usually does not exist in the backpropagation or weight update phases because the ubiquitous batch normalization destroys layer sparsity in the back-propagated gradient $\frac{\partial\!\mathcal{L}}{\partial y}$, so the additional hardware costs are not warranted for training. (We describe these challenges in more detail in \autoref{sec:background}.)

In this paper, we tackle the challenges of accelerating sparse training by combining algorithmic adaptation with dataflow and hardware optimizations. The accelerator architecture we propose, \ours{}, relies on four key insights:

\begin{enumerate}

\item Two-sided sparsity can only be leveraged in the forward pass, but increases interconnect complexity~\cite[etc.]{parashar2017scnn,chen2019eyerissv2}. \ours{} therefore exploits one source of sparsity in each training phase: weight sparsity in the forward and backward passes, and activation sparsity in the weight update phase. This maximizes energy and latency improvements while minimizing hardware complexity.

\item While load-balancing a sparse workload across a 2D PE array can destroy spatial reuse, spatial reuse generally arises in only one hardware dimension (either row or column broadcast). \ours{} uses dataflows that distribute the non-sparse minibatch dimension (always available during training) across one hardware dimension and the sparse tensor dimension(s) across the other hardware dimension, load-balancing the workload across the minibatch dimension. This achieves good utilization and preserves spatial reuse without a complex interconnect.

\item While sparse training approaches generally rely on sorting to determine which weights to keep, it actually suffices to \emph{partition} the weight set into two sets (retained and discarded). \ours{} replaces the sorting with a partitioning scheme based on dynamic quantile estimation~\cite{yazidi2017multiplicative}, which avoids the computation and storage overheads of sorting.

\item Weight initialization values are only important during early phases of training and quickly outweighed by the accumulated gradients. Therefore, in sparse training algorithms where retained initial weight components prevent computation sparsity~\cite{golub2019dropback}, the initial weights can be decayed to zero early in the training process.
\end{enumerate}

\noindent In the remainder of the paper, we first show how to adapt an existing sparse training algorithm~\cite{golub2019dropback} to make it suitable for hardware accelerator implementation; the adapted algorithm achieves $3.9\times$--$11.7\times$ sparsity while maintaining unpruned accuracy on tasks like CIFAR10 and ImageNet.

We then propose a hardware architecture that adapts a standard 2D-PE-array inference accelerator to enable sparse training without incurring the dataflow limitations and interconnect complexity of the only prior sparse training accelerator proposal~\cite{zhang2019eager} and achieves much higher sparsity. Finally, we develop a sparse data representation suitable for training access patterns, and an inexpensive load-balancing technique that preserves maximum spatio-temporal reuse without complicating the on-chip interconnect. Most of the modifications are not specific to the sparse training method we adapt, but rather are necessary for accelerating any existing sparse training approach.

Compared to an equivalent accelerator that does not support training-time sparsity, \ours{} uses $2.27\times$--$3.26\times$ less energy and offers $2.28\times$--$4\times$ speedup without compromising accuracy on state-of-the art networks on ImageNet and CIFAR-10.

\section{Sparse training considerations}
\label{sec:background}

\subsection{DNN training}
\label{sec:background:training}
\noindent Stochastic gradient descent (SGD) --- the de facto standard training algorithm for deep neural networks~\cite{lecun2012efficient} --- comprises three stages, illustrated in \autoref{fig:training-stages}:

\begin{enumerate}
\item The \emph{forward pass} runs the inference algorithm to determine the model's predictions for training inputs and calculate the loss $\mathcal{L}$ (i.e., the training error). For a convolutional layer, this consists of convolving the input activation (iact) tensor $x$ with a set of filters $w$ to obtain the output activation (oact) tensor $y$ (\autoref{fig:training-stages}a).
\item The \emph{backward pass} back-propagates the loss gradient across the model's layers. For a convolutional layer, this is done by convolving the loss gradient with respect to the oacts $\frac{\partial\!\mathcal{L}}{\partial y}$ with filters $w$; unlike in the forward pass, however, each filter is first rotated $180\degree$ (\autoref{fig:training-stages}b).
\item The \emph{weight update pass} determines how much a weight $w$ should be adjusted to decrease the loss by computing the gradient $\frac{\partial\!\mathcal{L}}{\partial w}$. For a convolutional layer, this consists of convolving the backpropagated loss gradient with respect to the oacts $\frac{\partial\!\mathcal{L}}{\partial y}$ with the input activations (iacts) $x$ (\autoref{fig:training-stages}c).
\end{enumerate}
\noindent In fully connected layers, $x$ and $y$ are 1D vectors, a weight matrix replaces the weight filters, inner product replaces convolution, and matrix transpose replaces the $180\degree$ rotation.

\subsection{Sources of sparsity}
\noindent Inference accelerators that support sparsity~\cite[etc.]{han2016eie,zhang2016cambriconx,parashar2017scnn,chen2019eyerissv2,gondimalla2019sparten} can leverage two sparsity sources: (a)~zero-valued weights that result from pruning~\cite{han2015learning}, and (b)~zero-valued activations that result from the \textsc{relu} activation function~\cite{albericio2016cnvlutin}. With suitable hardware support, multiply-accumulate (MAC) operations that involve zero weights or activations can be skipped, while zero-valued weights and activations need not be stored if a suitable sparse data format is used; some accelerators can take advantage of both sparsity sources simultaneously~\cite{parashar2017scnn,chen2019eyerissv2,gondimalla2019sparten}.

During training, weight sparsity can also be used in the backward gradient propagation phase, and input activation sparsity in the weight update phase (cf.~\autoref{fig:training-stages}). However, the back-propagated gradient $\frac{\partial\!\mathcal{L}}{\partial y}$ does not exhibit sparsity because of the prevalent use of batch normalization~\cite{ioffe2015batch}: batch normalization layers are commonly used between \conv{} and \textsc{relu} layers, which means that the $\frac{\partial\!\mathcal{L}}{\partial y}$ sparsity generated from backpropagating through \textsc{relu} is destroyed by backpropagating through the batch normalization layer.

Designers of sparse training accelerators, therefore, are faced with a choice: either spend additional hardware to accelerate one third of the training process, or reduce hardware complexity but give up on leveraging activation sparsity in the forward pass. In this paper, we focus on the latter approach.

\begin{algorithm}[t]
\renewcommand{\algorithmcfname}{Alg.}
\makeatletter
\renewcommand{\algocf@capseparator}{:}
\makeatother
\SetAlCapSty{}
\DontPrintSemicolon
\For(\hfill\it $\triangleright$ minibatch){$n \in \left[0,N\right)$}{
\For(\hfill\it $\triangleright$ filter x-dim){$r \in \left[0,R\right)$}{
\For(\hfill\it $\triangleright$ filter y-dim){$s \in \left[0,S\right)$}{
\For(\hfill\it $\triangleright$ oacts x-dim){$p \in \left[0,P\right)$}{
\For(\hfill\it $\triangleright$ oacts y-dim){$q \in \left[0,Q\right)$}{
\For(\hfill\it $\triangleright$ in channel){$c \in \left[0,C\right)$}{
\For(\hfill\it $\triangleright$ out channel){$k \in \left[0,K\right)$}{
$y[p, q, k, n] \text{~+=~} w[r, s, k, c]$
\strut~~~~~~~$\times\,x[p\!+\!r,q\!+\!s,c,n]$\;
}}}}}}}
\caption{The computation of a \conv{} layer forward pass.}
\label{alg:dataflow}
\end{algorithm}
\subsection{Mappings, dataflows, and load balancing}
\label{sec:dataflowbg}

\begin{figure}[t]
\begin{minipage}{0.39\columnwidth}
\includegraphics{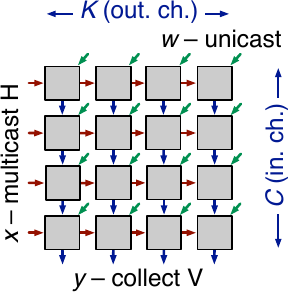}
\end{minipage}
\begin{minipage}{0.55\columnwidth}
\begin{tabular}{llll}
\toprule
&fw & bw & wu \\
\midrule
(H)orizontal & $x$ & $\sfrac{\partial\!\mathcal{L}}{\partial x}$ & $x$ \\
(V)ertical  & $y$ & $\sfrac{\partial\!\mathcal{L}}{\partial y}$ & $\sfrac{\partial\!\mathcal{L}}{\partial y}$ \\
(U)nicast & $w$ & $w$ & $\sfrac{\partial\!\mathcal{L}}{\partial w}$\\
\bottomrule\\[-0.6ex]
\multicolumn{4}{l}{$w$~weights~~~~~$x$~input activations}\\
\multicolumn{4}{l}{$y$~output activation partial sums}\\
\end{tabular}
\end{minipage}
\caption{A weight-stationary mapping: input and output channel dimensions ($C$ and $K$) are distributed spatially.}
\label{fig:ck-dataflow}
\end{figure}

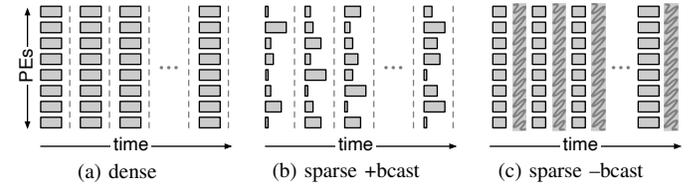
\begin{figure}[t]
\centering
\begin{tikzpicture}
\footnotesize
\node (img) {\includegraphics[width=\columnwidth]{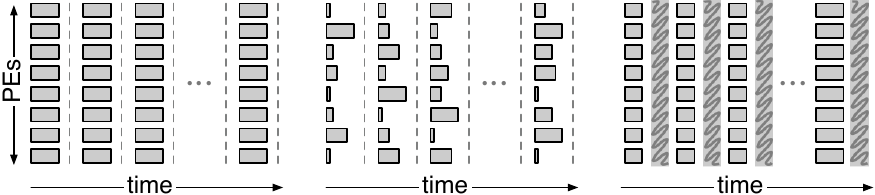}};
\node at (-88pt,-36pt){(a)~dense};
\node at (-1.5pt,-36pt){(b)~sparse +bcast};
\node at (83.5pt,-36pt){(c)~sparse --bcast};
\end{tikzpicture}
\vspace{-4ex}
\caption{DNN computation on a 2D PE array with a weight-stationary $C,K$ mapping: (a) dense model, equal work and spatial reuse; (b) sparse model, unequal work but spatial iact/psum reuse; (c) sparse model, equal work but no spatial iact/psum reuse. \protect\includegraphics{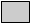} = work tile; \protect\includegraphics{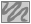} = overhead due to lack of reuse / complex interconnect. Each column is a full PE array's worth of work; a single layer's computation comprises many of these. $\pm$bcast = with/without spatial weight reuse.}
\label{fig:sparse:cks}
\vspace{-1ex}
\end{figure}

\noindent \autoref{alg:dataflow} illustrates the operations required to evaluate a \conv{} layer (the forward pass is shown but the backward and weight update passes can be expressed in the same way, with only the innermost MAC operation different). The computation can be represented as a seven-dimensional nested loop, where each loop traverses a different dimension of the operation space~\cite{parashar2019timeloop} (single-sample inference accelerators may not have the $N$ minibatch dimension). Regions of this operation space are then distributed as ``work tiles'' to different PEs by mapping two of the loops to the horizontal and vertical dimensions of a 2D PE array; together with exchanging the order in which loops are nested, this determines the dataflow~\cite{parashar2019timeloop,kwon2019understanding}.

\autoref{fig:ck-dataflow} shows the ubiquitous weight-stationary dataflow~\cite[etc.]{chen2014diannao,zhang2015optimizing,alwani2016fused,shen2016overcoming,suda2016throughput,jouppi2017datacenter,shen2017maximizing,nvdla2017}, which results from mapping the $C,K$ dimensions across the PE array in the forward pass (mapping $R,S$ is less common due to small filter sizes); the corresponding mappings for the backward and weight-update passes are shown in the adjacent table.

In this mapping, each workload (e.g., DNN layer) is first divided into PE-sized work tiles, all of which have the same number of weights. The tiles are mapped among the PEs; once PE receives one work-tile, the computation begins and runs until \emph{all} work-tiles have finished. Finally, the next set of work-tiles is distributed among the PEs, and the process repeats until the entire layer has been evaluated (\autoref{fig:sparse:cks}a).

This mapping results in advantageous dataflow properties in a 2D PE array. Because all work tiles have the same amount of work, execution is naturally synchronized, and data can be spatially reused by broadcasting across multiple PEs. For example, in the forward pass in \autoref{fig:ck-dataflow}, input activations are broadcast horizontally (read-only reuse), while partial sums are reduced vertically (read-write reuse). The dataflow patterns also allow the on-chip network to be simple: our example requires two one-dimensional flows (for the activations and the partial sums) and one unicast flow (for the weights); typically, those would be three separate interconnects.

However, difficulties arise when the network is sparse. At reasonable pruning levels, on the order of 10\% of the weights survive~\cite{han2015learning}, with sparsity distributed unevenly among the worktiles (by chance and learning pressure). This leaves designers with two unpleasant alternatives:

\begin{figure}
\includegraphics{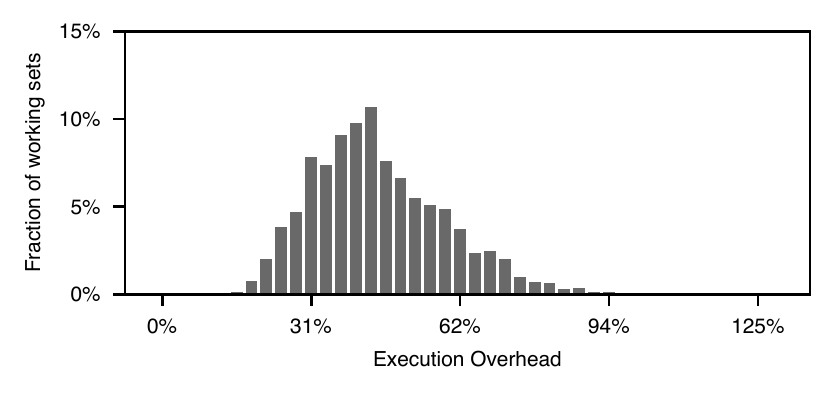}
\vspace{-2ex}
\caption{Load imbalance histogram of full-PE-array working sets (columns in \autoref{fig:sparse:cks}b) when training VGG-S~\cite{zagoruyko_torch_2015}/CIFAR-10~\cite{Krizhevsky2009LearningML} using Dropback sparse training~\cite{golub2019dropback}. A perfectly load-balanced workload would have 100\% of the sets at 0\% overhead. }
\label{fig:loadimb}
\vspace{-2ex}
\end{figure}

\begin{enumerate}
\item Retain the same tiling and mapping of operations to the PE array as in the dense case. This preserves the single-dimensional dataflow patterns shown in \autoref{fig:ck-dataflow}, allowing input activations and partial sums to be spatially reused. However, different amounts of work are distributed to different PEs, and utilization is low because latency is limited by the ``slowest'' PE (\autoref{fig:sparse:cks}b). \autoref{fig:loadimb} shows how latency differs among full-PE-array sets of work tiles (i.e., columns in \autoref{fig:sparse:cks}b): frequently, the load imbalance causes execution time overheads in excess of 50\%, and sometimes in excess of 100\%.

\item Distribute an equal number of non-zero weights to each PEs. This balances the workload among the PEs (\autoref{fig:sparse:cks}b), but destroys the desirable single-dimensional on-chip traffic flow patterns  of \autoref{fig:ck-dataflow} and severely reduces the benefits from spatial reuse. In addition, because related partial sums can be generated in any PE, a complex interconnect is required to reduce them~\cite{zhang2019eager,chen2019eyerissv2}.
\end{enumerate}

\noindent Choosing other dataflows also does not provide a panacea: for example, the activation-stationary dataflow used for some sparse accelerators~\cite{parashar2017scnn} suffers similar issues in the weight update pass, requires two datatypes to be unicast, and suffers from low PE array utilization towards the tail of many networks where the activation tensors are small~\cite{chen2019eyerissv2}.

\autoref{sec:loadbalancedataflow} describes how \ours{} employs the additional minibatch dimension available during training to achieve effective load balancing while preserving a hardware-friendly dataflow and avoiding the need for a complex interconnect.

\subsection{Sparse weight representation}
\label{sec:sparsewtbg}

\noindent Existing sparse-weight inference accelerators~\cite{han2016eie,zhang2016cambriconx,parashar2017scnn,chen2019eyerissv2,gondimalla2019sparten} employ a linear run-length encoding that is tightly coupled to the dataflow they use. For example, EIE~\cite{han2016eie} stores non-zero entries as an interleaved compressed sparse column (CSC) format, which permits a single column of an \fc{} layer weight matrix $W$ to be streamed to the PE array to interact with the same input activations. This layout matches the dataflow during the forward pass, but makes it impossible to calculate addresses within a column of $W^\top$ in the backward pass.

Similarly, the compressed format used for \conv{} filters in SCNN~\cite{parashar2017scnn} organizes filter layers so that all sparse filters with the same input channel (and different output channels) are adjacent. In the forward pass, this corresponds to SCNN's input-stationary dataflow where a single input activation is multiplied by all filters from the same input channel and the partial sums are distributed to different output channels; however, in the backward pass the equivalent gradient-stationary dataflow would need to compute addresses for all filters from one \emph{output} channel, which is not possible due to varying filter sparsity. 

\ours{} instead uses a variant compressed block format~\cite{csb} (\autoref{sec:rep}) to ensure that weights can be compressed but still read efficiently during all relevant training phases.

\subsection{Sparse training algorithms}
\label{sec:sparsetrain}

\noindent Training of sparse networks relies on the observation that dense deep neural networks contain small subnetworks ($\sim$20\% weights) that can be trained to match or exceed the original accuracy \emph{provided} that the initial weight settings for the subnetwork are retained~\cite{li2018measuring,frankle2019lottery}. In effect, most ($\sim$80\%--90\%) of the weights serve as a scaffolding necessary only to identify the weights that should survive in the final pruned subnetwork.

Most of the proposed sparse training algorithms work by gradually increasing sparsity during the training process. The lottery ticket algorithm~\cite{frankle2019lottery} prunes 20\% of the network every 50,000 training iterations by removing the lowest-magnitude weights; the authors report 5--10$\times$ model size reduction on CIFAR10 targets. Eager Pruning~\cite{zhang2019eager} follows a similar magnitude-based approach, but adds a feedback loop and a checkpoint-based rollback scheme to avoid overpruning; maintaining top-1 accuracy on ImageNet, it can prune ResNet50 2.4$\times$ (25.6M$\rightarrow$10.8M weights) by removing $0.8\%$ of the weights every 24,000 iterations. Both approaches rely on sorting all weight values to select which weights to keep.

Dynamic sparse reparametrization~\cite{mostafa2019parameter} starts by randomly distributing zero weights at the desired sparsity level, but allows the zeros to redistribute across the weight tensor during training. For ResNet50, for example, $\sim$200,000 additional parameters are set to zero every 1,000--8,000 iterations, but an equal number of weights are allowed to regrow after each pruning step. It avoids the need to sort all weights by using a value threshold adjusted via a set-point feedback loop whenever the network is pruned; however, the initial value of this threshold becomes a hyperparameter. ResNet50 can be pruned 3.5$\times$ (25.6M$\rightarrow$7.3M) with some top-1 accuracy loss on ImageNet ($-1.6\%$).

In contrast to the gradual pruning approaches~\cite{frankle2019lottery,zhang2019eager,mostafa2019parameter}, the Dropback algorithm~\cite{golub2019dropback} prunes the network from the beginning: only a fixed percentage of the parameters (e.g., 10\%) are ever allowed to change. In every iteration, only the weights with the highest accumulated gradient survive (which again requires sorting), on the theory that this represents learning better than magnitude during early iterations; the pruned weights are reset to their initial values rather than to 0. Dropback prunes ResNet18 11.7$\times$ (11.7M$\rightarrow$1M) while maintaining top-1 accuracy on ImageNet. 

In this paper, we focus on Dropback algorithm (\autoref{alg:dropback}), which offers by far the highest compression ratios and introduces only one additional parameter (the sparsity factor) during training. Unfortunately, two aspects stand in the way of hardware acceleration: (a)~pruned weights are not set to 0, and so MAC energy is not saved; and (b)~millions of gradients must be sorted to determine which weights should be pruned. We demonstrate how to overcome these drawbacks and make Dropback algorithm hardware-friendly in \autoref{sec:adaptalgo}.

\begin{algorithm}[t]
\renewcommand{\algorithmcfname}{Alg.}
\makeatletter
\renewcommand{\algocf@capseparator}{:}
\makeatother
\SetAlCapSty{}
\DontPrintSemicolon
\textbf{init:} $W^{(0)} \text{~with~} W^{(0)} \sim{} N\!\left(0, \sigma\right)$\;
\textbf{output:} $W^{(t)}$\;
	\While{not converged}  {
		$T = \left\lbrace\Big|\sum_{i=0}^{t-1} \frac{\eta \partial{}f\left(W^{(i-1)}; x^{(i-1)}\right)}{\partial w} \Big| \text{~s.t.~} w \in W_\mathit{trk} \right\rbrace$\;
		$P = \left\lbrace\Big|\frac{\eta \partial{}f\left(W^{(i-1)}; x^{(i-1)}\right)}{\partial w} \Big| \text{~s.t.~} w \in W_\mathit{prn} \right\rbrace$\;
		$S = \operatorname{sort}\!\left(T \cup P\right)$\;
		$\mathit{mask} = \mathbbm{1}\!\left(S > S\left[k\right]\right)$\;
		$W^{(t)} = \mathit{mask} \,\kern-1pt\cdot{}\kern-1pt \left(W^{(t-1)}\!-\!\eta \nabla\!f\!\left(W^{(t-1)}; x^{(t-1)}\right)\right) + \overline{\mathit{mask}} \cdot{} W^{(0)}$\;
		$t = t+1$\;
	}
\caption{Dropback algorithm \cite{golub2019dropback}. $W_\mathit{trk}$ and $W_\mathit{prn}$ = tracked and pruned weights; $T$ and $P$ = tracked and pruned accumulated gradients; $S$ = sorted accumulated gradients; $k$ = number of gradients to keep; $\eta$ = learning rate. $\mathit{mask}$ is a boolean matrix indicating which weights to keep and $\overline{\mathit{mask}}$ is its logical inverse.\vspace{-1ex}}
\label{alg:dropback}
\end{algorithm}

\section{Adapting sparse training algorithms to hardware}
\label{sec:adaptalgo}

\noindent To adapt Dropback algorithm to the requirements of an efficient hardware implementation, \ours{}

\begin{enumerate}
\item[(i)] \label{enum:dblim:rst} creates computation sparsity by decaying initial weight values $W^{(0)}$ over the first 1,000 iterations, and
\item[(ii)] \label{enum:dblim:sort} avoids the need to sort all gradients by using dynamic quantile estimation to continuously determine a threshold value that tracks the target sparsity.
\end{enumerate}
We discuss the details below.

\begin{figure}[t]
\includegraphics{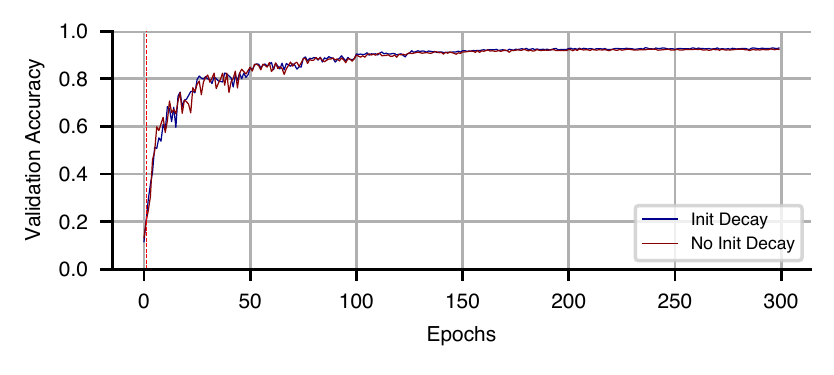}
\vspace{-2ex}
\caption{Validation accuracy over the course of training when initial weights decay 0.9$\times$ every iteration, compared to a baseline without decay (VGG-S on CIFAR-10). The dashed vertical line indicates the point at which all initial weights have decayed to zero (1,000 iterations, or early in the second epoch as epochs are 800 iterations each).}
\label{fig:decayaccuracy}
\vspace{-2ex}
\end{figure}

\subsection{Creating computation sparsity}
\label{sec:creatingsparsity}
\noindent A key challenge in using Dropback algorithm~\cite{golub2019dropback} to enable energy-efficient training is the fact that it never entirely removes pruned weights: instead, pruned weights have their values returned to their initialization-time values. These are generally non-zero, so no MAC operations are saved, and, because MAC computation accounts for much of the energy during training (cf.~Figures~\ref{fig:opportunity} and~\ref{fig:results:energy}), energy savings are also limited.

To determine how to recover computation sparsity, we first considered the function of the weights during training. We hypothesized that the initial weight values are important during the early iterations when weights have not moved far from their initial state and the accumulated gradients are small compared to the initial weights. Later on, we reasoned, the accumulated gradients are much larger than the initial weight values, and the initial scaffolding could safely be removed.

We therefore examined whether the initial weight values could be gradually decayed to zero so that eventually only the accumulated gradients remain and all pruned weights become zero. We decayed the initial weight values 10\% every iteration (decay parameter $\lambda=0.9$), eventually zeroing them; the resulting training scheme is detailed in \autoref{alg:dropbackdecay}. \autoref{fig:decayaccuracy} shows how validation accuracy evolves over the course of training compared to a baseline where weights do not decay: neither accuracy nor convergence time are affected.

\begin{algorithm}
\renewcommand{\algorithmcfname}{Alg.}
\makeatletter
\renewcommand{\algocf@capseparator}{:}
\makeatother
\DontPrintSemicolon
\SetAlCapSty{}
\textbf{init:} $W^{(0)} \text{~with~} W^{(0)} \sim{} N\!\left(0, \sigma\right)$\;
\textbf{output:} $W^{(t)}$\;
	\While{not converged}  {
		$T = \left\lbrace\Big|\sum_{i=0}^{t-1} \frac{\eta \partial{}f\left(W^{(i-1)}; x^{(i-1)}\right)}{\partial w} \Big| \text{~s.t.~} w \in W_\mathit{trk} \right\rbrace$\;
		$P = \left\lbrace\Big|\frac{\eta \partial{}f\left(W^{(i-1)}; x^{(i-1)}\right)}{\partial w} \Big| \text{~s.t.~} w \in W_\mathit{prn} \right\rbrace$\;
		$S = \operatorname{sort}\!\left(T \cup P\right)$\;
		$\mathit{mask} = \mathbbm{1}\!\left(S > S\left[k\right]\right)$\;
		$W^{(t)} = \mathit{mask} \,\kern-1pt\cdot{}\kern-1pt \left(W^{(t-1)}\!-\!\eta \nabla\!f\!\left(W^{(t-1)}; x^{(t-1)}\right)\right) + \overline{\mathit{mask}} \cdot{} \lambda^t W^{(0)}$\;
		$t = t+1$\;
	}
\caption{Dropback algorithm with initial weight decay. $W_\mathit{trk}$ and $W_\mathit{prn}$ = tracked and pruned weights; $T$ and $P$ = tracked and pruned accumulated gradients; $S$ = sorted accumulated gradients; $k$ = number of gradients to keep; $\eta$ = learning rate; $\lambda$ = decay parameter (we used 0.9). $\mathit{mask}$ is a boolean matrix indicating which weights to keep and $\overline{\mathit{mask}}$ is its logical inverse.}
\label{alg:dropbackdecay}
\end{algorithm}

\noindent In this experiment, the initial weight decay scheme results in 80\% weights set to zero by iteration 1000 (out of 234,400 total iterations). This means that 60\% of computation in 99.5\% of iterations can be entirely skipped, potentially resulting in significant energy savings.

\subsection{Choosing which weights to keep}
\noindent The second key challenge of the original Dropback algorithm is the need to sort all accumulated gradients to determine which weights should be kept and which should be reset to their initial values. A comparison-based sort requires a minimum of $\log_2\left(n!\right)$ comparisons in the worst case --- 336M comparisons for the relatively compact VGG-S with 15M weights, compared to the 4.3G MACs required for one training iteration with batch 16. Even if the DNN accelerator were modified to support sorting (i.e., to return both indices and values), sorting would take in excess of 1.3M cycles on a 256-PE device.

To overcome this challenge, we considered replacing the target sparsity factor (such as 10$\times$) with a global value threshold $\vartheta$. In this scheme, every computed gradient is tested whether it should be added to the tracked set $T$, and added to $T$ only if it exceeds $\vartheta$. This would reduce the number of comparisons to one per produced gradient (15M for VGG-S).

The question is how to determine $\vartheta$ for each iteration. Dynamic sparse reparametrization~\cite{mostafa2019parameter} accomplishes this via a set-point feedback scheme that adjusts $\vartheta$ every 1,000--8,000 iterations, but this introduces an additional hyperparameter, the initial value of $\vartheta$. Instead, we determine $\vartheta$ dynamically via a streaming quantile estimation technique~\cite{yazidi2017multiplicative}, shown in \autoref{alg:quantile}. To allow for peak update rate (up to 4 per cycle in the last VGG-S \conv{} layer), we extended the technique to process four updates at once. 

The tracking process proceeds as follows:
\begin{itemize}
\item If the gradient dimension $\delta_w$ is \emph{not} in the tracked set $T$, $|\delta_w|$ is compared against $\vartheta$. If it is higher, $\delta_w$ evicts and replaces the lowest entry in $T$; otherwise, it is discarded. In either case, $|\delta_w|$ is used to update the quantile estimate (\autoref{alg:quantile}).
\item If $\delta_w$ is tracked, it is added to the stored accumulated gradient $\delta_w^\mathrm{acc}$. The quantile estimate is updated with $|\delta_w^\mathrm{acc}+\delta_w|$.
\end{itemize}
In our experiments, we found that the tracking accuracy sensitivity to the values of $\widehat{Q}_q(0)$ and $\varrho$ is negligible, so we use the same values for all experiments (see \autoref{alg:quantile}) rather than treating them as hyperparameters.
 
\begin{algorithm}[t]
\renewcommand{\algorithmcfname}{Alg.}
\makeatletter
\renewcommand{\algocf@capseparator}{:}
\makeatother
\SetAlCapSty{}
\DontPrintSemicolon
\textbf{init:} $\widehat{Q}_q(0) = 10^{-6}; \varrho = 10^{-3}$\;
\textbf{input:} $\delta(n), \widehat{Q}_q(n)$\;
\textbf{output:} $\widehat{Q}_q(n+1)$\;
\If{$\widehat{Q}_q(n) < \delta(n)$}{
$\widehat{Q}_q(n+1) = \left(1+\varrho q\right) \widehat{Q}_q(n+1)$
}\Else{
$\widehat{Q}_q(n+1) = \left(1-\varrho\left(1- q\right)\right) \widehat{Q}_q(n+1)$
}
\caption{The quantile estimation algorithm DUMIQUE~\cite{yazidi2017multiplicative}. $\delta(n)$ = the $n^\mathrm{th}$ accumulated gradient value computed; $\widehat{Q}_q(n)$ = the $q^\mathrm{th}$ quantile estimate at step $n$; $\varrho$ = adjustment rate hyperparameter. \ours{} uses a modified, parallelized variant which treats the average of four incoming accumulated gradients as a single $\delta(n)$.}
\label{alg:quantile}
\end{algorithm}

\begin{figure}[t]
\includegraphics{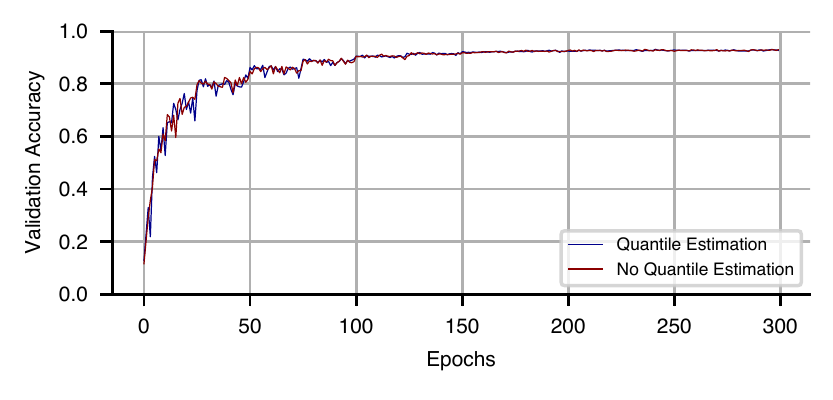}
\vspace{-2ex}
\caption{Validation accuracy over training epochs when sparse training is used and quantile estimation (\autoref{alg:quantile}) is used to determine the value threshold $\vartheta$ under which accumulated gradients are discarded, compared to a baseline with initial weight decay and exact sorting (VGG-S on CIFAR-10).}
\label{fig:qeaccuracy}
\vspace{-2ex}
\end{figure}

To determine the accuracy of this estimate, we trained VGG-S using a sparsity target of $7.5\times$ and streamed the computed accumulated gradients to the estimator. \autoref{fig:qeaccuracy} shows that while the quantile estimation exhibits minor deviations from ground truth (because different layers have different amounts of sparsity), these estimation errors have no detrimental effect on the validation accuracy of the trained network. Overall, the quantile estimation error results in extra weights being tracked, and reduces the sparsity factor slightly from $7.5\times$ to $5.2\times$; however, this overhead is much lower than that required to sort all weights or to train a dense network.

Note that selecting weights through quantile estimation is not specific to the Dropback algorithm: separating some fraction of the highest-value or highest-gradient weights is needed by all sparse training algorithms~\cite{mocanu2018scalable,golub2019dropback,frankle2019lottery,mostafa2019parameter}.

\section{Dataflow \& sparse data format}

\subsection{Storage and sparsity during training}

\noindent Weights (or, more precisely, accumulated gradients) are always stored compressed using the format described in \autoref{sec:rep}. Typically, all weight gradients are produced, but most gradients that are not already tracked will not survive the comparison with existing accumulated gradients.

Activations are stored uncompressed for immediate reuse and in a compressed format for long-term reuse. The forward pass reads sparse weight tensor, and produces a dense output activation tensor, which is then immediately reused as inputs to the next layer; the activations are then compressed using a sparse, zero-free format, and reused in the weight update stage. This technique is similar in spirit to Gist~\cite{jain2018gist}.

\subsection{Compressed sparse weight representation}
\label{sec:rep}

\begin{figure}[t]
\includegraphics{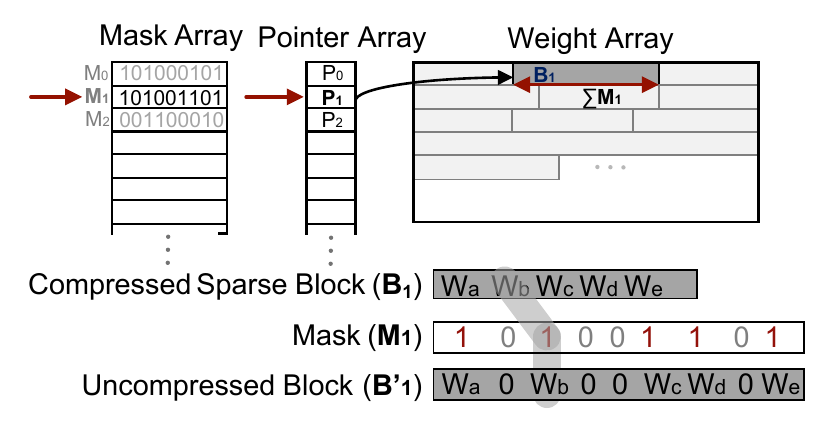}
\vspace{-0.5em}
\caption{The compressed sparse block (CSB) weight representation in \ours{}.}
\label{fig:sparserep}
\vspace{-2ex}
\end{figure}

\noindent To avoid the challenges discussed in \autoref{sec:sparsewtbg}, a sparse weight storage format designed for training must support:
\begin{enumerate}
\item[(i)] \label{enum:sparsereqiter} iterating through 2D convolution filters across different dimensions in different stages (for \conv{} layers), and across both rows and columns of weight matrices (for \fc{} layers),
\item[(ii)] \label{enum:sparsereqrot} rotating kernels (for \conv{} layers) or transposing weight matrices (for \fc{} layers), and
\item[(iii)] \label{enum:sparsereqsize} different kernel sizes in \conv{} layers.

\end{enumerate}

\noindent \ours{} uses a modified compressed sparse block (CSB) format~\cite{csb} shown in \autoref{fig:sparserep} to store weights in the on-chip global buffer and external DRAM. Blocks store non-zero values and are variable in size because of sparsity, but correspond to fixed-size regions in the corresponding dense weight space --- kernels for \conv{} layers, square fragments of the weight matrix in \fc{} layers, etc. The region size can vary on layer granularity to support different kernel sizes.

The \ours{} CSB format comprises three components, illustrated in \autoref{fig:sparserep}:
\begin{enumerate}
\item[(a)] the weight array, which stores variable-size packed weight blocks corresponding to kernels, etc.;
\item[(b)] the pointer array, indexed by tensor coordinates, which identifies the weight array location that stores the relevant weight values; and
\item[(c)] the mask array, also indexed by tensor coordinates, which stores a mask identifying non-zero value locations in the unpacked block (and therefore also the packed size).
\end{enumerate}
The pointer and mask arrays are decoupled to support different mask lengths for each layer (e.g., different kernel sizes in \conv{} layers, flexible block sizes in \fc{} layers and during weight update, and so on); in all of our simulations, mask arrays fit in the on-chip GLB.

Because the pointer array is indexed by coordinates in the original (dense) operation space and is decoupled from the compressed contents, the format makes computing kernel addresses straightforward while adapting cleanly to different kernel dimensions. The indirection also makes it easy to determine the density of working sets assigned to each PE: it suffices to subtracting pointers of adjacent work tiles. In addition, because blocks are sized to and retrieved on filter granularity, they can be rotated (to be used in the backprop pass) while being fetched from the global buffer to the per-PE register files; similarly, transposition of the weight matrix for the \textsc{fc} layers can be done by transposing subtensors piecewise.

Activations are stored uncompressed for short term reuse (as activations in the next layer) and compressed in CSB format for long-term reuse (forward pass to weight update).

\subsection{Load balancing and dataflow}
\label{sec:loadbalancedataflow}

\begin{figure}[t]
\centering
~\includegraphics{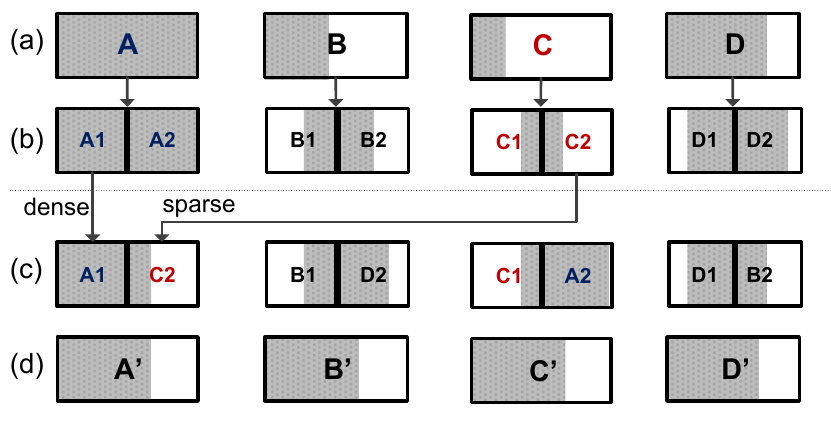}
\vspace{-1ex}
\caption{Load balancing: work tiles are cut in half~(b) and the halves rearranged in dense-sparse pairs~(c).}
\label{fig:loadbalance}
\vspace{-2ex}
\end{figure}

\noindent \autoref{fig:loadbalance} illustrates the load balancing process used in \ours{}. First, every work tile~(a) is cut into two halves along one of the tile dimensions~(b); because sparsity is almost certainly uneven within the tile, the two halves will likely have different densities. Next, the halves are sorted according to density, and half-tiles are matched starting from opposite ends~(c): the sparsest half-tile is matched with the densest half-tile, and so on. This ensures that each newly formed tile is as close as possible to the average density across all PE work tiles~(d).

\begin{figure}[t]
\includegraphics[width=\columnwidth]{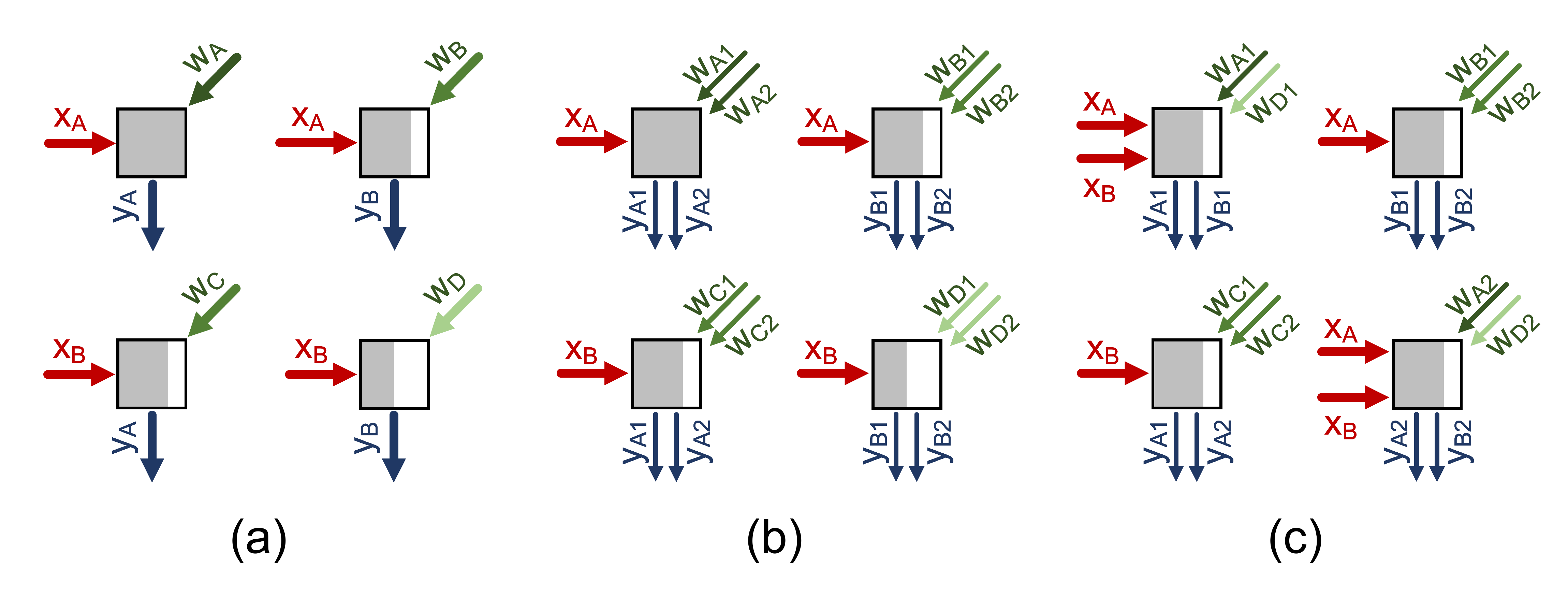}
\vspace{-4ex}
\caption{Load balancing in the weight-stationary $C,K$ dataflow in a four-PE array: (a)~PE workload imbalance (shaded PEs) due to different weight sparsities (shaded arrows); (b)~PE workloads and the corresponding weight ($w$) and partial sum ($y$) tiles split in half across the $K$ dimension (note the thinner arrows); (c)~half-tiles exchanged between the top-left and bottom-right PEs for load balancing. Activations must be sent on both rows and columns, and require twice the buffer space in the PEs.}
\label{fig:lb:ck}
\end{figure}

\begin{figure}[t]
\begin{minipage}{0.39\columnwidth}
\includegraphics{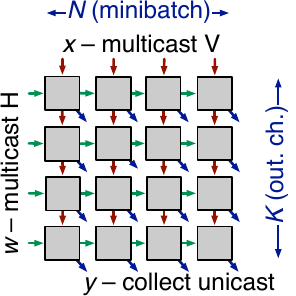}
\end{minipage}
\begin{minipage}{0.55\columnwidth}
\begin{tabular}{llll}
\toprule
&fw & bw & wu \\
\midrule
(H)orizontal & $w$ & $w$ & $\sfrac{\partial\!\mathcal{L}}{\partial w}$ \\
(V)ertical  & $x$ & $\sfrac{\partial\!\mathcal{L}}{\partial x}$ & $x$ \\
(U)nicast & $y$ & $\sfrac{\partial\!\mathcal{L}}{\partial y}$ & $\sfrac{\partial\!\mathcal{L}}{\partial y}$\\
\bottomrule\\[-0.6ex]
\multicolumn{4}{l}{$w$~weights~~~~~$x$~input activations}\\
\multicolumn{4}{l}{$y$~output activation partial sums}\\
\end{tabular}
\end{minipage}
\caption{Mappings and dataflows that spatially distribute the minibatch across one dimension of the PE array.}
\label{fig:kn-dataflow}
\vspace{-2ex}
\end{figure}

\begin{figure}[t]
\includegraphics[width=\columnwidth]{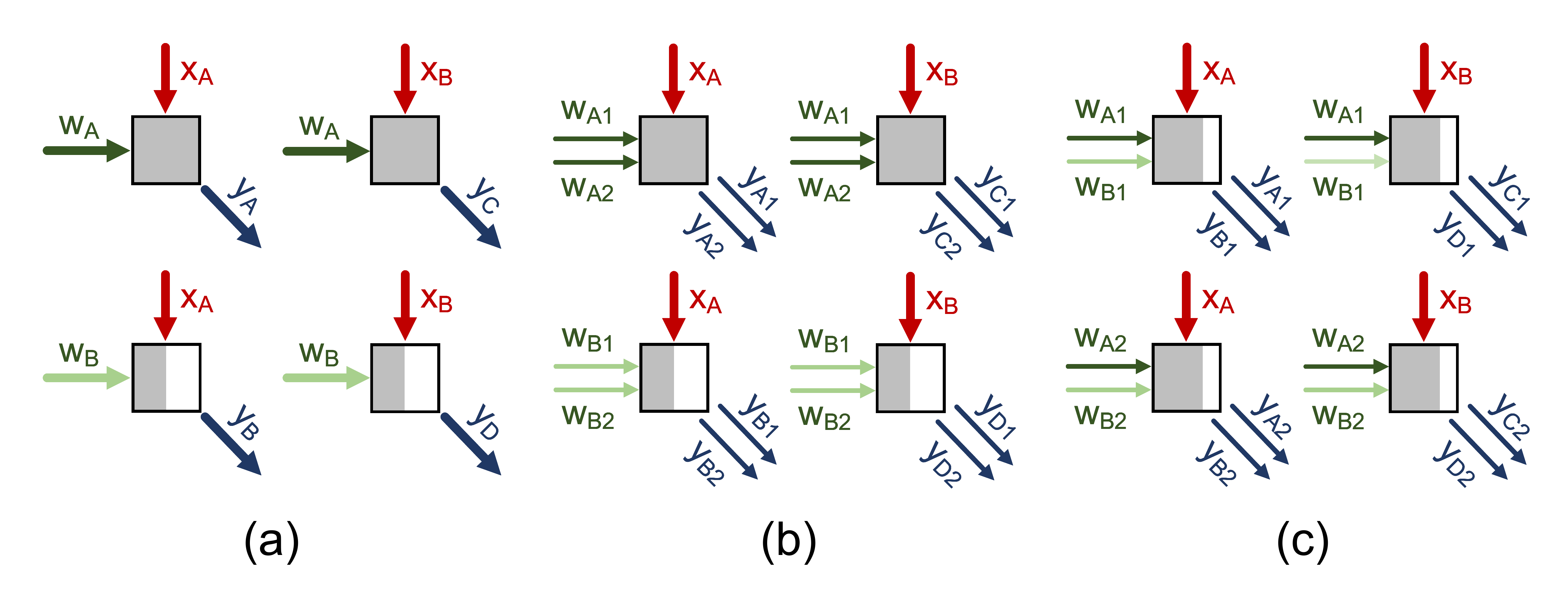}
\vspace{-4ex}
\caption{Load balancing in the proposed $K,N$ dataflow in a four-PE array: (a)~PE workload imbalance (shaded PEs) due to different weight sparsities (shaded arrows); (b)~PE workloads and the corresponding weight ($w$) and partial sum ($y$) tiles split in half across the $K$ dimension (note the thinner arrows); (c)~half-tiles exchanged between the top-left and bottom-right PEs to load-balance across $K$. Each input activation tile is still sent to only one column.}
\label{fig:lb:kn}
\end{figure}

\begin{figure}[t]
\includegraphics{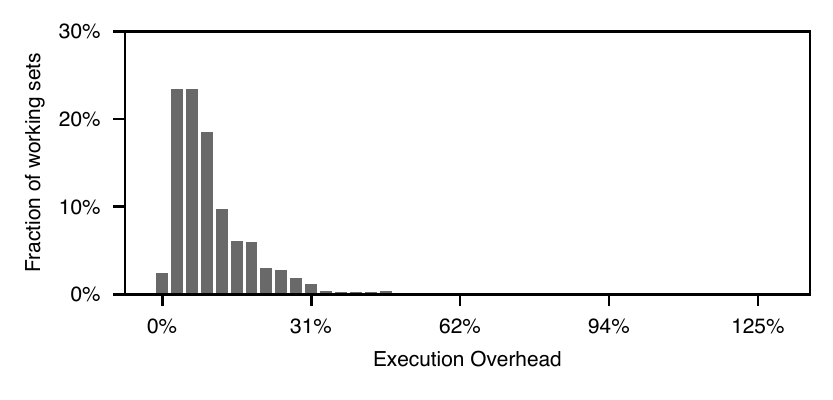}
\vspace{-2ex}
\caption{Load imbalance histogram of full-chip working sets (columns in \autoref{fig:sparse:cks}b) after load-balancing half-tiles (VGG-S/CIFAR-10). Compare with \autoref{fig:loadimb}. A perfectly load-balanced workload would have 100\% of the sets at 0\% overhead.}
\label{fig:loadimbfixed}
\vspace{-1ex}
\end{figure}

However, naively applying this rebalancing scheme to the entire PE array without changing the dataflow would impact on-chip communications patterns and require a complex interconnect. \autoref{fig:lb:ck} demonstrates this on the weight-stationary $C,K$ dataflow on a 4-PE array. In pane~(a), input activations are broadcast horizontally ($x_A$ in the top row and $x_B$ in the bottom row), partial sums are accumulated vertically ($y_A$ in the left column and $y_B$ in the right column), while the weights are unicast (as in~\autoref{fig:ck-dataflow}); however, because the weights have different levels of sparsity (shaded arrows),  the PEs have different amount of computation (shaded PEs). In pane~(b), each PE's workload is cut in half as discussed above; each weights tile ($w_A$ and $w_B$) is also split in half (e.g., into $w_{A1}+w_{A2}$ and $w_{B1}+w_{B2}$, note the thinner arrows), as are the corresponding partial sums ($y_A$ and $y_B$). Finally, in pane~(c), the workload halves are balanced across the PE array, so that the top-left and bottom-right PEs swap half their workloads; this, however, means that all input activations ($x_A$ and $x_B$) must now be sent to both columns and rows, requiring more bandwidth and a more complex interconnect, and double the activations must be buffered at the target PEs. The $P,Q$ input-stationary dataflow faces similar challenges in the weight update pass (cf.~\autoref{fig:training-stages}) and requires unicasting two of the three datatypes.

\ours{} addresses both of these problems by leveraging a simple observation: training is typically done across a minibatch of 32--64 samples rather than on single items~\cite{bengio2012practical,masters2018revisiting}.\footnote{Minibatches in the 1,000s allow faster training on large multi-GPU clusters but can incur some accuracy cost~\cite{goyal2017accurate,akiba2017extremely}.}

Because a training accelerator does not need to support single-sample inference, the minibatch dimension ($N$ in \autoref{alg:dataflow}) can be used to distribute work tiles across one dimension of the PE array. The other dimension can then be safely chosen to be a dimension where sparsity exists --- e.g., the input or output channel dimensions ($C$ or $K$) with weight sparsity. Because only one dimension is sparse, and that dimension corresponds to spatial reuse, the load balancing process needs to be applied only to one dimension of the PE array (i.e., the dimension opposite to the spatial reuse pattern, here $N$).

\autoref{fig:kn-dataflow} illustrates how a $K,N$ mapping (output channel, minibatch) with load balancing across the output channel ($K$) dimension preserves the single-dimension dataflow properties (cf.~\autoref{fig:ck-dataflow}) during the forward pass. Weights are now the same across the minibatch and are multicast across the horizontal dimension of the PE array, partial sums are collected across the vertical dimension, and input activations vary across both dimensions and so are unicast.

A detailed example is shown in \autoref{fig:lb:kn}. As in \autoref{fig:lb:ck}, pane~(a) shows the unbalanced workload, pane~(b) shows each PE's workload (and consequently the weight and partial sum tiles) cut into half, and pane~(c) shows the PE array after load-balancing along the $K$ dimension. Observe that, in contrast to \autoref{fig:lb:ck}, the load-balanced dataflow in pane~(c) has the same on-chip interconnect communication patterns and requires the same interconnect bandwidth as the unbalanced dataflow in pane~(c).

Finally, \autoref{fig:loadimbfixed} demonstrates that this technique results in effective load balancing. While the balance is not 100\% perfect, the execution time overheads for most full-PE-array working sets are small at <10\%, with the worst imbalance at 30\% --- a vast improvement to the common 40\%--50\% overheads and up to $2\times$ slowdown without load balancing (see \autoref{fig:loadimb}).

\section{Hardware architecture}
\label{sec:arch}
\noindent The overall hardware architecture of \ours{} is based on 2D PE array where each PE has a local register file~(RF) and all PEs share an on-chip global buffer~(GLB); an off-chip DRAM completes the memory hierarchy. PEs are interconnected via three simple interconnects: two support one-dimensional traffic flows in the horizontal and vertical directions, and one supports unicast traffic to any PE in the array. Because \ours{} focuses on training, we use 32-bit floating point MAC units in the PE datapath, but the design can be used with any datatype.

\begin{figure}[t]
\centering
\includegraphics[width=0.9\columnwidth]{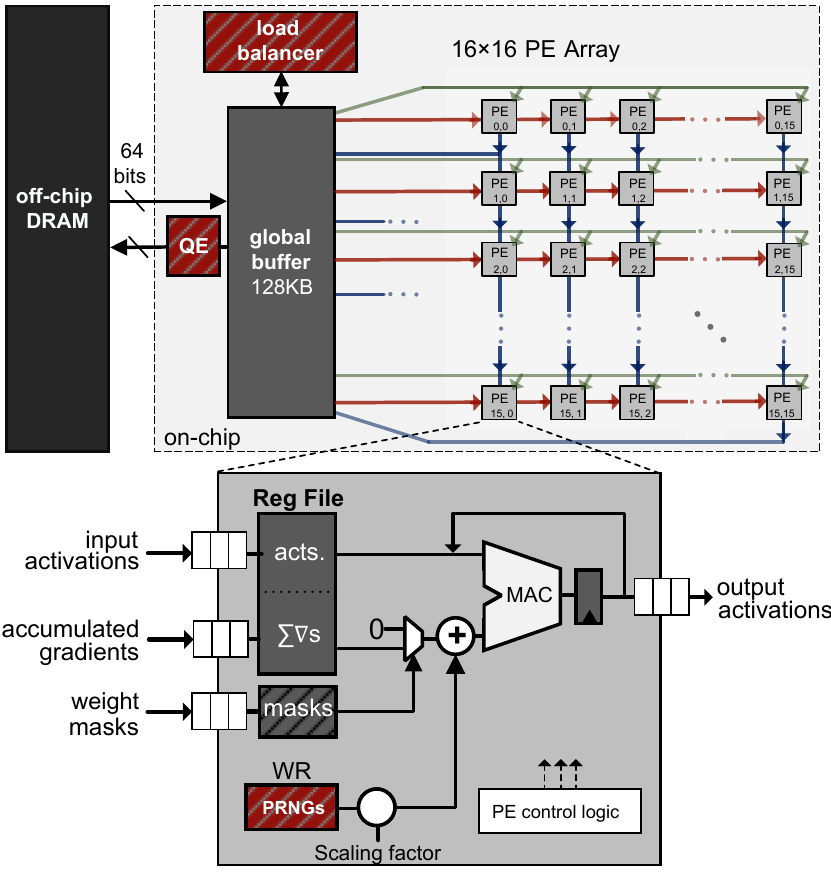}
\vspace{-1ex}
\caption{\ours{} system architecture. The WR module is added to recreate initial weights for Dropback-style training, the QE unit is added to support quantile estimation, and the load balancer is added to support work tile re-balancing.}
\label{fig:arch}
\end{figure}

The design is illustrated in \autoref{fig:arch}, with differences from the baseline accelerator dashed. \ours{} places one global quantile estimation unit (QE) between global buffer and the external DRAM; the QE unit monitors accumulated gradients flowing from the GLB to DRAM and discards all except those above the target sparsity quantile.

In addition, each PE contains a weight recomputation unit (WR) responsible for generating the initial weight values. The WR accepts a weight index and generates a 32-bit integer initial value for the relevant weight. It consists of 3 xorshift~\cite{marsaglia2003xorshift} pseudo-random generators (RNGs) whose outputs are added to produce an approximately Gaussian output. Note that, unlike conventional RNG, the WR unit does not contain hidden state, and is purely a function of its seed and the weight index. The ``RNG'' output is then scaled using an integer multiplier; this this enables popular initialization formul\ae{} like Xavier~\cite{glorot2010understanding} or Kaiming~\cite{he2015delving}, and allows the initial weights to be decayed as per \autoref{alg:dropbackdecay}. Finally, the scaled value is converted to FP32 and added to the accumulated gradient retrieved from weight storage if the weight is tracked, or to zero if the weight has been pruned.

\begin{table}
\begin{center}
\sf\small
\begin{tabular}{ll}
\toprule
\multicolumn{2}{c}{\bfseries Baseline dense accelerator} \\ 
\midrule
PEs & 256 (16$\times$16) \\
datatype & 32-bit floating-point \\
pruning type & none \\
interconnect & 3$\times$ 1D-flow interconnect\\
global buffer & 128 KB \\
local buffer (RF) & 1 KB per PE \\
dataflow & optimal (via Timeloop+Accelergy) \\
\toprule
\multicolumn{2}{c}{\bfseries \ours{} modifications} \\ 
\midrule
pruning type & lowest accumulated gradients \\
pseudo-RNG & xorshift~\cite{marsaglia2003xorshift}, one per PE \\
quantile estimator & DUMIQUE~\cite{yazidi2017multiplicative}, max 4 requests / cycle \\
dataflow & optimal spatial-minibatch dimension \\
\bottomrule
\end{tabular}
\end{center}
\vspace{-2ex}
\caption{Hardware configurations for the baseline dense training accelerator and \ours{} sparse training accelerator.}
\label{tbl:config}
\vspace{-4ex}
\end{table}

\begin{table*}[t]
	\begin{center}
	\sf\small
	\begin{tabular}{llllllllll}
	\toprule
	model & dataset & dense  & dense  & sparse  & sparse & sparsity & \# epochs & dense  & pruned  \\ 
	 &  &  size & MACs & size &  MACs &  & &  accuracy & accuracy \\ 
	\midrule
	Densenet & CIFAR-10 & 2.7M & 528M & 692k & 157M & 3.9$\times$ & 340 & 94.2\% & 93.7\% \\
	WRN-28-10 & CIFAR-10 & 36M & 4G & 8.3M & 863M & 4.3$\times$ & 462 & 96.0\% & 96.1\% \\
	VGG-S & CIFAR-10 & 15M & 269M & 2.9M & 113M & 5.2$\times$ & 236 & 93.0\% & 93.1\% \\
	\midrule
	MobileNet~v2 & ImageNet & 3.5M &  301M & 0.35M & 75M & 10$\times$ & 131 & 70.98\% & 71.13\% \\
	ResNet18 & ImageNet & 11.7M & 1.8G & 1M & 359M & 11.7$\times$ & 81 & 69.17\% & 69.31\% \\
	\bottomrule
	\end{tabular}
	\end{center}
\vspace{-2ex}
	\caption{Sparsity achieved using the \ours{} training scheme for the CNNs tested, together with weight footprint and MAC reduction and the final accuracy compared to the dense baseline.}
	\label{tbl:sparsity}
	\end{table*}
	
	\begin{figure*}
	\includegraphics{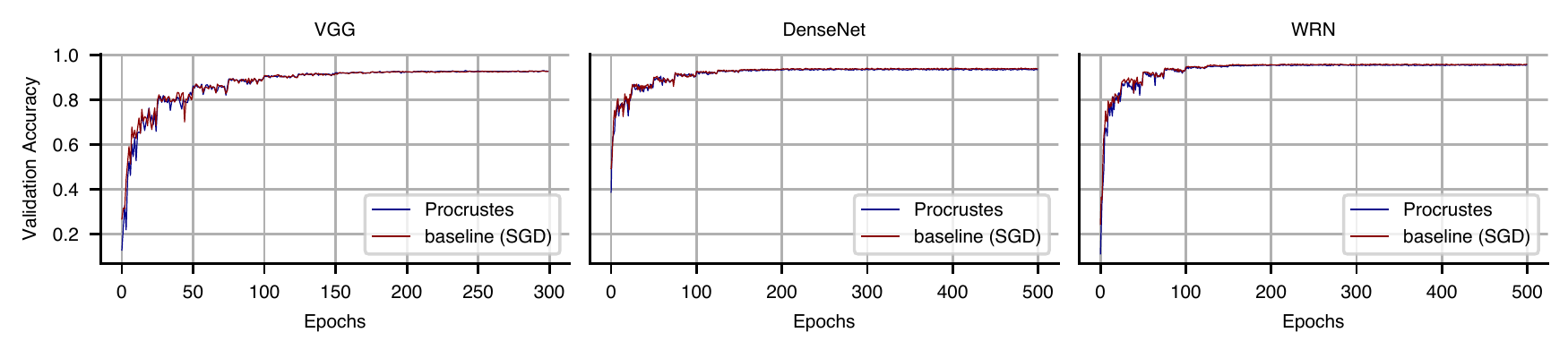}
	\vspace{-2ex}
	\caption{Validation accuracy over training time for \ours{} and the unpruned baseline (SGD) on CIFAR-10: (left) VGG-S, (centre) DenseNet, and (right) WRN-10-28.}
	\label{fig:results:acc}
	\end{figure*}
	\begin{figure*}
	\centering
	\includegraphics[width=0.29\textwidth]{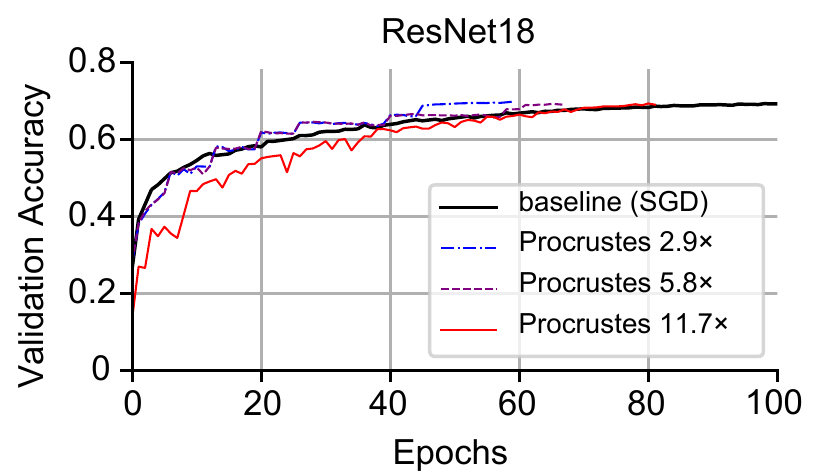}
	\qquad\qquad
	\includegraphics[width=0.29\textwidth]{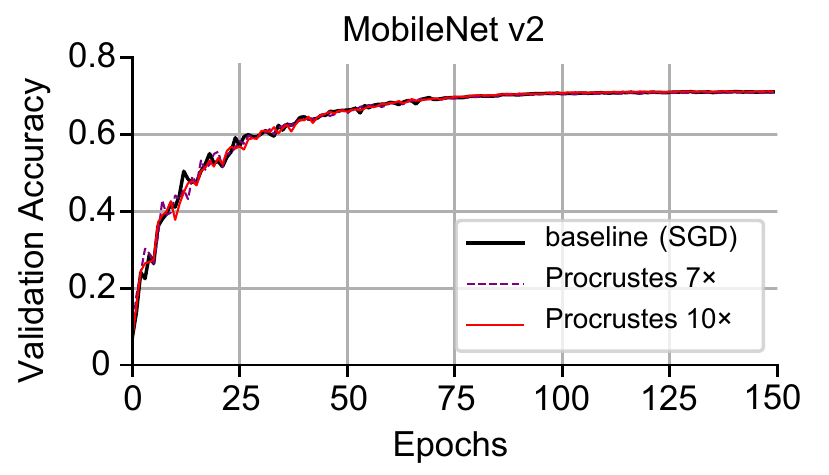}
	\caption{Validation accuracy over training time for Procrustes and the unpruned baseline for ResNet18 (left) and MobileNet~v2 (right) on ImageNet.} 
	\label{fig:results:acc-imnet}
\vspace{-2ex}
	\end{figure*}

\section{Evaluation}
\label{sec:evaluation}

\subsection{Methods}
\label{sec:methods}

\noindent We re-implemented the baseline Dropback training algorithm~\cite{golub2019dropback} using PyTorch~\cite{pytorch} and verified the reported training sparsity levels and accuracy results; we then implemented the initial weight decay and quantile-estimation extensions needed for \ours{}.

We evaluated \ours{} on five CNNs: ResNet18~\cite{hedeep2016} (11.7M weights) and MobileNet~v2 (3.5M weights) applied to the ImageNet image classification task~\cite{Russakovsky:ImageNet:2015}, as well VGG-S~\cite{zagoruyko_torch_2015} (a 9.2$\times$ reduced version of VGG-16 with 15M weights), WRN-28-10~\cite{zagoruyko2016wide} (36.5M weights), and a small Densenet~\cite{huang2017densely} (growth rate 24, 3 blocks $\times$ 10 layers, for a total of 2.7M weights), all CIFAR-10~\cite{Krizhevsky2009LearningML}.

To determine optimal mappings and dataflows, we extended Timeloop~\cite{parashar2019timeloop} to support sparse weight masks (retrieved from our PyTorch model), model sparse computation, account for sparse encoding overheads, and accurately reflect latency due to load imbalances. We also used Timeloop to determine cycle-level latency; to determine energy costs, we use energy access cost provided in Accelergy~\cite{wu2019accelergy} with its default 40nm library. We modelled all layers of all networks and all stages of training (forward, backward, and weight update).

As a dense baseline, we used a 2D PE array architecture with 16$\times$16 PEs, adapted to the 32-bit floating-point precision commonly used in training; we used Timeloop to determine the optimal tiling and dataflow. Hardware modules not present in the baseline were implemented in Verilog~RTL and synthesized using Synopsys~DC in the 45nm FreePDK process. Accelerator configuration details are shown in \autoref{tbl:config}.

\subsection{Pruning ratios and accuracy}
\noindent\autoref{tbl:sparsity} shows the sparsity factors achieved while maintaining the same accuracy as the corresponding dense (unpruned) network using the \ours{} sparse training algorithm. Depending on the network, our training scheme achieves $3.9\times$--$11.7\times$ weight sparsity without compromising accuracy.

Importantly, achieving unpruned-level accuracy does not require additional convergence time. 
\autoref{fig:results:acc} demonstrates this on the VGG-S, DenseNet, and WRN, all on CIFAR-10. \autoref{fig:results:acc-imnet} demonstrates the same effect on ResNet18 trained on ImageNet at various weight pruning ratios. Overall, \ours{} converges reaches state-of-the-art accuracy as quickly (or faster) than the baseline unpruned network.

\subsection{Energy savings and speedup}
\begin{figure*}
\includegraphics[width=\textwidth]{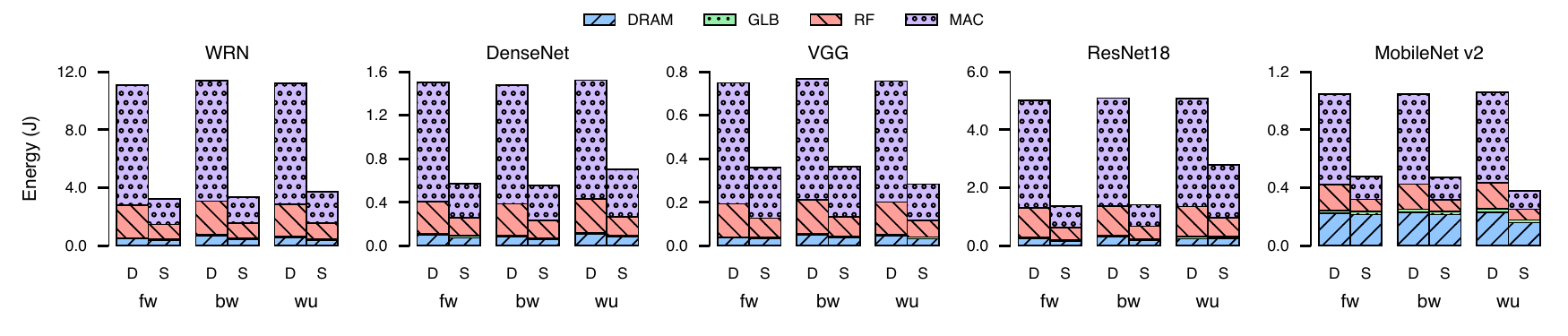}
\vspace{-4ex}
\caption{Energy breakdown of using KN dataflow for (left) WRN-10-28, (middle left) DenseNet, (middle) VGG-S, (middle right) ResNet18, and (right) MobileNet~v2. Lower is better. K = output channel dimension; N = minibatch dimension. S = sparse; D = dense. fw = forward pass; bw = backward pass; wu = weight update phase.}
\label{fig:results:energy}
\vspace{-2ex}
\end{figure*}

\noindent \autoref{fig:results:energy} shows the energy savings obtained by training with \ours{} across several CNNs. Most of the energy is saved by performing fewer MAC operations; because training is most often done on FP32 values, MACs dominate the energy usage. Intra-PE register file (RF), global buffer (GLB), and DRAM access energies are also substantially reduced, but account for less of the baseline energy expenditure, and therefore contribute less to the overall savings.

The figure also illustrates that \ours{} can transform higher sparsity ratios into bigger energy savings: ResNet18, which has the highest pruning factor (11$\times$), saves the most energy compared to the dense baseline (3.26$\times$), while WRN has the best speedup (4$\times$). MobileNet~v2 benefits less in energy because its depth-separable convolutions limit reuse and so comparatively more energy is spent on DRAM accesses; however, Procrustes still trains it with 2.39$\times$ less energy than the dense baseline, and almost as much speedup as WRN (3.88$\times$ faster than dense).

For most networks, the forward and back-propagation passes offer more energy savings; this is because those passes can take advantage of weight sparsity, which is generally higher than activation sparsity. VGG-S demonstrates a less common case where the weight sparsity is concentrated in the layers that perform relatively few MACs, so the activation sparsity leveraged by the weight-update phase actually saves more operations.

Overall, \ours{} is effective in converting training-time sparsity to energy savings.

\subsection{Mapping and dataflow choice}

\begin{figure*}
\includegraphics[width=\textwidth]{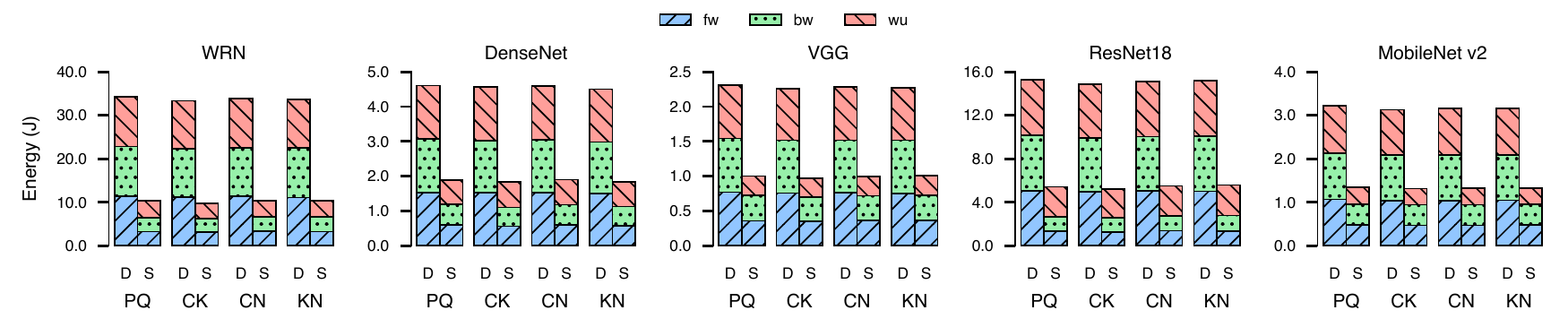}
\vspace{-4ex}
\caption{Energy Comparison across different dataflows for (left) WRN-10-28, (middle left) DenseNet, (middle) VGG-S, (middle right) ResNet18, and (right) MobileNet~v2. Lower is better. C = input channel dimension; K = output channel dimension; P and Q = output activation dimensions; N = minibatch dimension. S = sparse; D = dense. fw = forward pass; bw = backward pass; wu = weight update phase.}
\label{fig:results:energy_dataflow}
\vspace{-2ex}
\end{figure*}

\noindent \autoref{fig:results:energy_dataflow} shows how energy expenditure varies with different spatial partitioning schemes. Sparsity enables energy improvement across all phases and all mappings. Because the number of MAC operations and the memory hierarchy are the same across the different mappings, the lion's share of the energy use is the same across the different dataflows, and variations are negligible. This is in agreement with prior work that also reported negligible impact of the chosen dataflow on the energy during inference~\cite{yang2018overrated}.

This finding enables us to select spatial partitioning that results in the best performance (i.e., shortest execution time).

\begin{figure*}
\includegraphics[width=\textwidth]{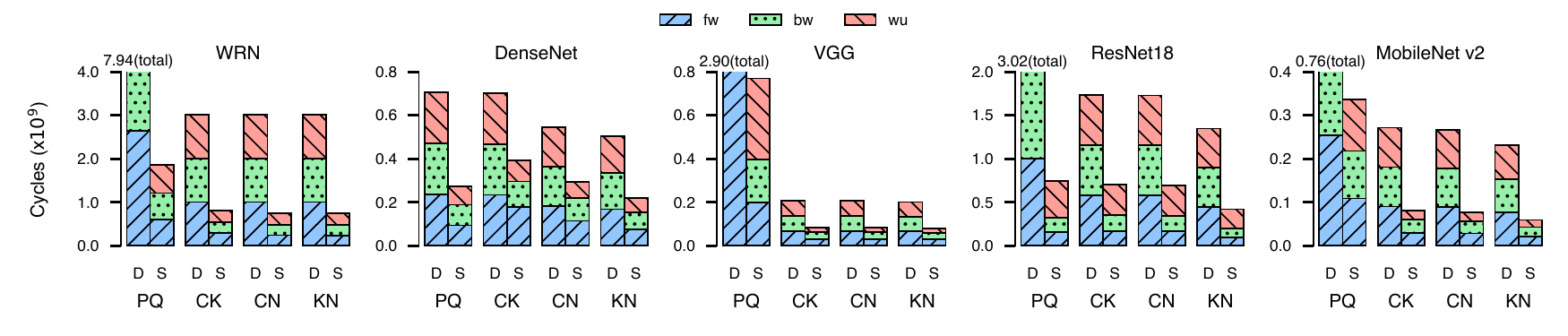}
\caption{Training latency across different dataflows for (left) WRN-10-28, (middle left) DenseNet, (middle) VGG-S, (middle right) ResNet18, and (right) MobileNet~v2. Lower is better.
 C = input channel dimension; K = output channel dimension; P and Q = output activation dimensions; N = minibatch dimension. S = sparse; D = dense. fw = forward pass; bw = backward pass; wu = weight update phase.}
\label{fig:results:speedup}
\end{figure*}

\autoref{fig:results:speedup} shows how execution times vary when the working set is mapped to the PE array using different spatial partitioning schemes; all schemes can be implemented using the simple network topology shown in \autoref{fig:arch} except the weight-stationary $C,K$ scheme, which requires a complex network to load-balance PE working sets across the entire chip. The partitioning schemes that distribute the minibatch dimension along one of the PE array dimensions ($C,N$ and $K,N$) are the fastest mappings because they are able to achieve effective load balancing and good utilization across all layers of the CNNs; $K,N$ performs slightly better because it offers slightly higher utilization in the first network layer. The $C,K$ scheme performs less well even though it requires a more complex interconnect, largely because it is inefficient on layers that have few channels. The activation-stationary $P,Q$ scheme does not require load-balancing in the forward and back-propagation phases, is hard to load-balance during the weight update phase, and has low utilization when activation tensors are relatively small; it is overall the slowest mapping.

\ours{} uses the overall fastest $K,N$ scheme for all phases of training.

\subsection{Scalability}
\begin{figure*}
	\centering
    \includegraphics{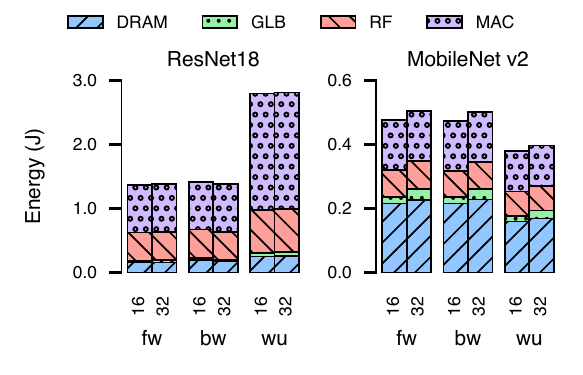}~\includegraphics{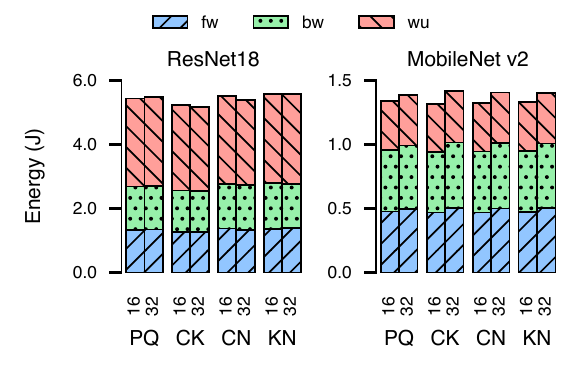}~\includegraphics{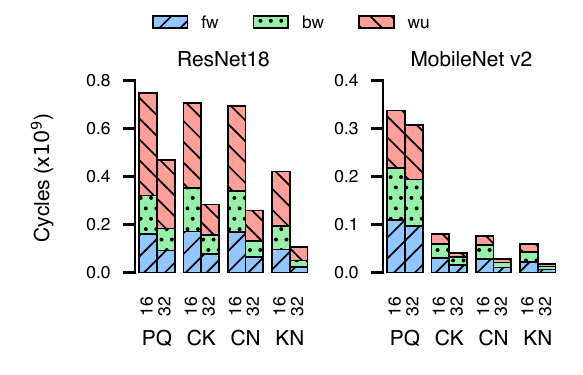}
    \vspace{-1em}
    \caption{Scalability of Procrustes on 16$\times$16 (256) to 32$\times$32 (1024) cores on ResNet-18 and MobileNet~v2 classifying ImageNet configured as in Figs.~\ref{fig:results:energy}--\ref{fig:results:speedup}. Energy differences are is negligible as the workload is the same. Speedup scales best for the Procrustes mappings (CN and KN) because other mappings trade off utilization for reuse.} 
    \label{fig:results:scalability}
    \vspace{-1em}
\end{figure*}

\noindent \autoref{fig:results:scalability} shows how Procrustes scales when the PE array size is quadrupled from 256 cores (16$\times$16) to 1024 cores (32$\times$32); the global buffer size is doubled over the 256-core size (a factor of $\sqrt{2}$). Overall, energy is very similar same for all dataflows / passes because the number of MAC operations is the same. Latency scales near ideally (3.9$\times$ on 4$\times$ the cores) in the $K,N$ mapping used by Procrustes. Other mappings (especially activation-stationary $P,Q$) do not scale as well since they trade off PE array utilization to retain spatial reuse.

\subsection{Silicon area overheads}

\noindent The silicon area and power overheads of \ours{} are detailed in \autoref{tbl:overheads}. Despite the RNG initial weight recomputation module being included in every PE, its area and power pale in comparison to the FP32 MAC unit which all PEs include.

Overall, the \ours{} accelerator has an area overhead of 14\% over an equivalent dense accelerator, and consumes 11\% more power when executing the same \emph{dense} workloads. Both are a small price to pay for the $2.27\times$--$3.26\times$ energy savings offered by sparse training.

\begin{table}
\begin{center}
\sf\small
\begin{tabular}{ccc}
\toprule
{\bfseries Component}  & {\bfseries Power} ($mW$) & {\bfseries Area} ($\mu m^2$) \\ 
\toprule
\multicolumn{3}{c}{\bfseries Per-PE area: \ours{} overheads \emph{italicized}} \\
\midrule
FP32~MAC            & 7.29  & 18,875.72   \\
Register File       & 15.61 & 198,004.71  \\
\emph{PRNG}                & \emph{0.35}  & \emph{1,920.84}    \\
\emph{Mask Memory}         & \emph{2.65}  & \emph{44,932.66}   \\
\toprule
\multicolumn{3}{c}{\bfseries System area: \ours{} overheads \emph{italicized}} \\ 
\midrule
Global Buffer       & 73.74 & 17,109,596.5 \\
\emph{Quantile Engine} & \emph{1.38}  & \emph{9,861.4}     \\
\emph{Load Balancer} & \emph{2.05}  & \emph{8,725.23}     \\
\bottomrule
\end{tabular}
\end{center}
\vspace{-2ex}
\caption{Silicon area costs and overheads (synthesis using Synopsys~DC with the FreePDK 45nm library). For fairness, the power estimates assume the same dense computation (i.e., no sparsity).}
\vspace{-2ex}
\label{tbl:overheads}
\end{table}

\subsection{Generality}
\noindent \ours{} is the first sparse training accelerator to combine substantial sparsity ratios, $2.27\times$--$3.26\times$ energy savings, and up to $4\times$ speedups while maintaining state-of-the-art accuracy of the trained networks. While in this paper we use \ours{} to extend the Dropback training algorithm, the quantile estimation and spatial-minibatch dataflow insights apply to all existing --- and likely many future --- sparse training algorithms.

\section{Related work}
\label{sec:related}

\subsection{Sparse accelerators}
\noindent Eager Pruning~\cite{zhang2019eager} is the only extant proposal for a sparse training accelerator. It works by starting with a dense network and very gradually pruning the lowest-magnitude weights, with fewer than 1\% of weights removed every tens of thousands of training iterations; maintaining accuracy limits pruning to comparatively low factors of 1.5--3.5$\times$.
The accelerator uses on a weight-stationary dataflow where denser filters are distributed over more PEs than sparser filters; to manage the resulting irregularity in collecting partial sums, the authors propose a module that connects the PEs and can either accumulate or route partial sums. Although the Eager Pruning algorithm relies on sorting weights, this does not appear to be considered in the hardware or the latency and energy measurements. In contrast, \ours{} achieves higher pruning factors, does not rely on sorting weights, and avoids the need for a complex interconnect via a novel load balanced dataflow.

All other sparse accelerators only support inference. 
EIE~\cite{han2016eie} and CambriconX~\cite{zhang2016cambriconx} use a variants of the compressed sparse column format, which prevents them from efficiently accessing weights during the backward pass. SCNN~\cite{parashar2017scnn} and SparTen~\cite{gondimalla2019sparten} use an input-stationary dataflow to enable both weight and activation sparsity; however, both use a CSC-like format to encode sparse weights, and neither can be used to accelerate training.

\subsection{Sparse training algorithms}

\noindent Most proposed sparse training algorithms very slowly increase sparsity during the training process. The lottery ticket algorithm~\cite{frankle2019lottery} prunes 20\% of the network every 50,000 training iterations by removing the lowest-magnitude weights; the authors report 5--10$\times$ model size reduction on CIFAR10 targets. Eager Pruning~\cite{zhang2019eager} follows a similar magnitude-based approach, but adds a feedback loop and a checkpoint-based rollback scheme to avoid overpruning; maintaining top-1 accuracy on ImageNet, it can prune ResNet50 2.4$\times$ (25.6M$\rightarrow$10.8M weights) by removing $0.8\%$ of the weights every 24,000 iterations. Unlike \ours{}, both approaches rely on sorting all weight values to determine which weights to keep.

Dynamic sparse reparametrization~\cite{mostafa2019parameter} starts by randomly distributing zero weights at the desired sparsity level, but allows the zeros to redistribute across the weight tensor during training. For ResNet50, for example, $\sim$200,000 additional parameters are set to zero every 1,000--8,000 iterations, but an equal number of weights are allowed to regrow after each pruning step. It avoids the need to sort all weights by using a value threshold adjusted via a set-point feedback loop whenever the network is pruned; however, the initial value of this threshold becomes a hyperparameter. This method prunes ResNet50 3.5$\times$ (25.6M$\rightarrow$7.3M) with some top-1 accuracy loss on ImageNet ($-1.6\%$). \ours{} offers higher sparsity factors and prunes the network much more quickly, which translates to substantial energy savings.

Dropback~\cite{golub2019dropback} prunes the network from the beginning: only a fixed percentage of the parameters (e.g., 10\%) are ever allowed to change. In every iteration, only the weights with the highest accumulated gradient survive (which again requires sorting), on the theory that this represents learning better than magnitude during early iterations; the pruned weights are reset to their initial values rather than to 0. With Dropback, ResNet18 can be pruned 11.7$\times$ (11.7M$\rightarrow$1M) while maintaining top-1 accuracy on ImageNet. \ours{} adapts Dropback to the needs of an efficient hardware implementation, removing the requirement for sorting and decaying initial weights to 0 to create computation sparsity.

\section{Summary}
\noindent This paper introduces \ours{}, a sparse DNN training accelerator that produces pruned models with the same accuracy as dense models without first training, then pruning, and finally retraining, a dense model.

\ours{} relies on three key techniques. First, it adapts an existing training algorithm to create computation sparsity that can be converted into energy savings. Next, it replaces the sorting step present in nearly all sparse training algorithms with hardware-friendly, computationally simple quantile estimation. Finally, it leverages a novel load-balancing scheme that converts sparsity into speedup, and proposes a novel dataflow that enables load balancing without significant changes to the on-chip interconnect.

\section{Acknowledgements}

\noindent The authors are grateful to the anonymous reviewers for insightful feedback and helpful suggestions.

This material is based on research sponsored by Air Force Research Laboratory (AFRL) and Defense Advanced Research Project Agency (DARPA) under agreement number FA8650-20-2-7007, and by the Natural Sciences and Engineering Research Council of Canada (NSERC) under award number NETGP~485577-15. The U.S. Government is authorized to reproduce and distribute reprints for Governmental purposes notwithstanding any copyright notation thereon. The views and conclusions contained herein are those of the authors and should not be interpreted as necessarily representing the official policies or endorsements, either expressed or implied, of Air Force Research Laboratory (AFRL), Defense Advanced Research Project Agency (DARPA), the U.S. Government, the Natural Sciences and Engineering Research Council of Canada (NSERC), or the Government of Canada.

\bibliographystyle{IEEEtranS}
\fancypagestyle{plain}{}
\bibliography{IEEEabrv,sparse-train-accel-micro2020}

\end{document}